\documentclass[lettersize,journal]{IEEEtran}
\usepackage{amsmath,amsfonts}
\usepackage{algorithmic}
\usepackage{algorithm}
\usepackage{array}
\usepackage[caption=false,font=normalsize,labelfont=sf,textfont=sf]{subfig}
\usepackage{textcomp}
\usepackage{stfloats}
\usepackage{url}
\usepackage{verbatim}
\usepackage{graphicx}
\usepackage{cite}
\usepackage{hyperref}       
\usepackage{url}            
\usepackage{booktabs}       
\usepackage{amsfonts}       
\usepackage{nicefrac}       
\usepackage{microtype}      
\usepackage{xcolor}         
\usepackage{amsmath}
\usepackage{amssymb}
\usepackage{mathtools}
\usepackage{amsthm}
\usepackage{multirow}
\usepackage{wrapfig}

\begin{document}

\title{Contrastive Local Manifold Learning for No-Reference Image Quality Assessment}

\author{{Zihao Huang, Runze Hu, Timin Gao, Yan Zhang, Yunhang Shen, Ke Li}
\thanks{Zihao Huang and Runze Hu are with the School of Information
and Electronics, Beijing Institute of Technology, Beijing 100000, China
(e-mail: 3120240675@bit.edu.cn; hrzlpk2015@gmail.com).}
\thanks{Timin Gao and Yan Zhang are with the School of Key Laboratory of Multimedia Trusted Perception and Efficient Computing, Xiamen University, Xiamen 361005, China (e-mail: timingao@stu.xmu.edu.cn; bzhy986@gmail.com).}
\thanks{Yunhang Shen and Ke Li are with YouTu Laboratory, Tencent Company, Shanghai 518064, China (e-mail: shenyunhang01@gmail.com; tristanli@tencent.com).}
\thanks{Corresponding author: Yan Zhang.}}

\markboth{Journal of \LaTeX\ Class Files,~Vol.~14, No.~8, August~2021}%
{Shell \MakeLowercase{\textit{et al.}}: A Sample Article Using IEEEtran.cls for IEEE Journals}

\IEEEpubid{}

\maketitle

\begin{abstract}

Image Quality Assessment (IQA) methods typically overlook local manifold structures, leading to compromised discriminative capabilities in perceptual quality evaluation. To address this limitation, we present LML-IQA, an innovative no-reference IQA (NR-IQA) approach that leverages a combination of local manifold learning and contrastive learning. Our approach first extracts multiple patches from each image and identifies the most visually salient region. This salient patch serves as a positive sample for contrastive learning, while other patches from the same image are treated as intra-class negatives to preserve local distinctiveness. Patches from different images also act as inter-class negatives to enhance feature separation. Additionally, we introduce a mutual learning strategy to improve the model’s ability to recognize and prioritize visually important regions.  Comprehensive experiments across eight benchmark datasets demonstrate significant performance gains over state-of-the-art methods, achieving a PLCC of \textbf{0.942} on TID2013 (compared to 0.908) and \textbf{0.977} on CSIQ (compared to 0.965). 

\end{abstract}

\begin{IEEEkeywords}
Image Quality Assessment, Local Manifold Learning.
\end{IEEEkeywords}

\section{Introduction}
\label{sec:intro}

Image Quality Assessment (IQA) is critical in computer vision, aiming to replicate human visual perception in quality evaluations. Unlike simplistic pixel-based evaluations, IQA encompasses sophisticated perceptual factors, e.g., edge sharpness, color fidelity, and contrast, which are central to human visual judgments. The significance of accurate IQA has grown notably with the proliferation of digital media, where high-quality visual content is crucial for numerous applications~\cite{banham1997digital, dong2015image, li2025zooming}, such as autonomous driving, multimedia streaming, and virtual reality. Among the three main categories of IQA methods, i.e., Full-Reference (FR), Reduced-Reference (RR), and No-Reference (NR), NR-IQA has attracted substantial research interest due to its practicality and challenging nature, as it evaluates image quality without any reference information.

Deep learning has significantly advanced the capabilities of IQA models~\cite{ke2021musiq,TReS,DEIQT,li2025distilling}. However, several notable limitations persist. A key challenge is the limited availability of annotated data for supervised learning, primarily due to the prohibitive costs associated with subjective image quality evaluations. To overcome this issue, current approaches typically employ transfer learning. This involves using feature representations extracted from models that were pre-trained on classification tasks, such as Convolutional Neural Networks (CNNs) or Vision Transformers (ViTs). Despite their utility, these approaches inherently prioritize semantic categorization rather than nuanced perceptual quality, which often neglects fine-grained visual distortions crucial to human quality assessment. Consequently, models adapted directly from classification tasks tend to inadequately represent subtle distortions that significantly impact perceived image quality.

To tackle these challenges, self-supervised learning has emerged as a promising alternative~\cite{saha2023re,madhusudana2022image}, enabling models to learn from local image patches without requiring explicit labels. These methods typically define crops from the same image as positives while treating crops from different images as negatives. However, these approaches typically do not consider that perceived quality is often non-uniform across an image. This oversight can consequently lead to inaccuracies in the final quality evaluation.

\begin{figure*}[t]
  \centering
    \includegraphics[width=0.8\textwidth]{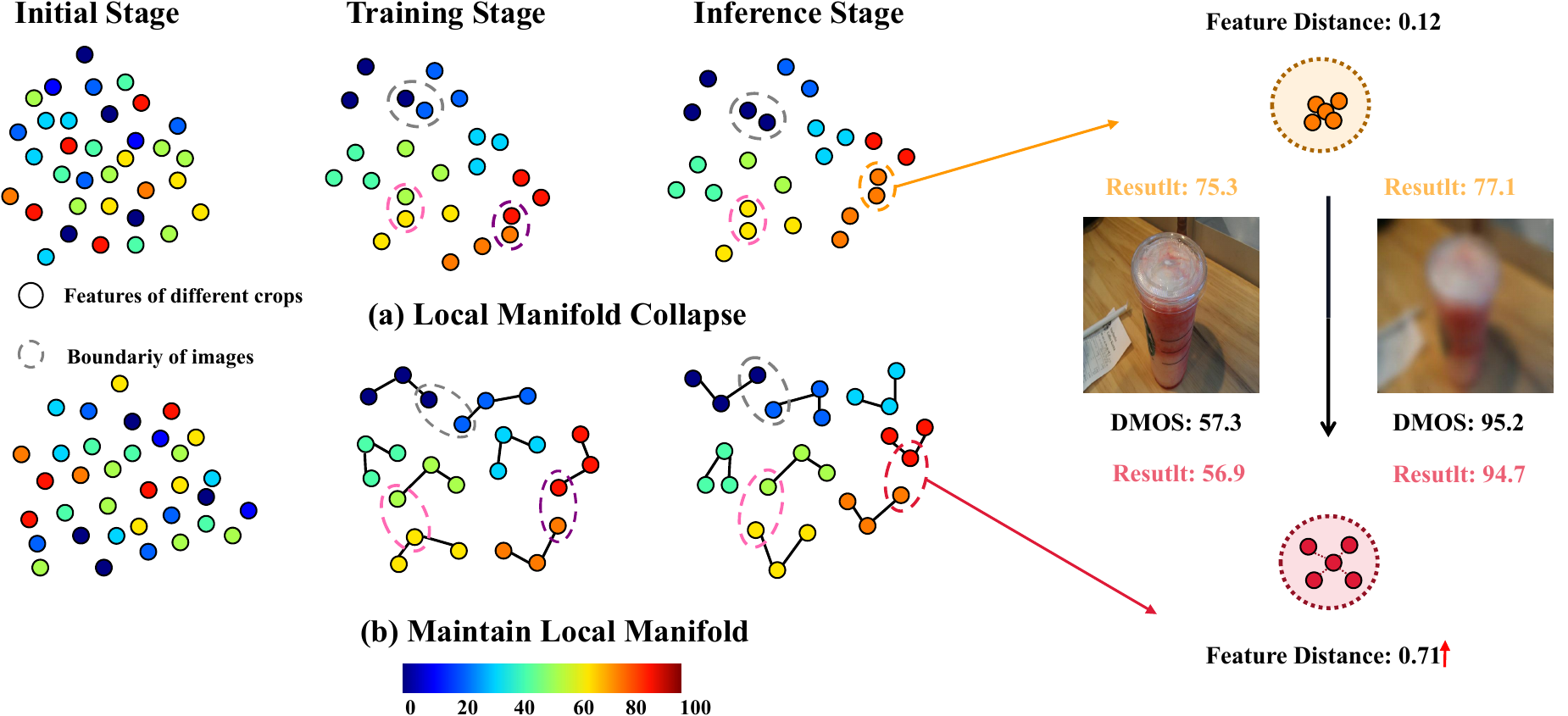}
  \caption{
  Illustration of contrastive learning paradigms in NR-IQA, with MOS values represented by circle color. (a) Prior IQA methods using contrastive learning aim for feature convergence among all crops of an image. The consequence is a failure to retain the local manifold, illustrated by the progressively smaller feature distances inside the dashed box. (b) Our framework is designed to preserve the local manifold, thereby upholding a diverse feature space and maintaining consistent feature distances.
}
  \label{fig1}
\end{figure*}

Real-world evidence suggests that image quality is inherently content-dependent and can be influenced by diverse viewer preferences~\cite{li2018has, sun2023blind}. This complexity manifests in the disparity between local patch scores and an image's overall quality rating~\cite{ying2020patches, zhang2018blind}. Additionally, cropping-based data augmentation can cause local patches from the same image to become excessively similar in feature space, leading to local manifold collapse. As depicted in Fig.~\ref{fig1}, such a collapse disrupts the global manifold structure, compromising the reliability of quality predictions. Specifically, when two images contain identical semantic content but exhibit different distortion levels, their local manifolds may become indistinguishable, reducing their feature space separation in terms of $L1$ distance. This convergence can obscure crucial local quality differences, resulting in imprecise quality predictions. Thus, in IQA, contrastive learning should differentiate images and account for quality variations across different regions within the same image. Maintaining sufficient intra-class separation is essential to prevent local manifold collapse and ensure more accurate and robust quality assessments. 
%
%
%


To address these limitations, we propose a innovative contrastive learning framework LML-IQA. This framework leverages Local Manifold Learning (LML) to enhance feature manifold differentiation not only across different images but also within the same image.
We introduce a mutual learning strategy consisting of teacher and student models. The teacher model operates by extracting visually salient crops, which is a critical step in evaluating overall image quality. In parallel, the student model works to refine local manifold structures.

Specifically, the teacher model selects image regions with the highest attention scores as positive samples. In contrast, other patches within the same image and patches from different images serve as negative samples. This contrastive setup enables the model to learn local manifold representations effectively.
The student model’s encoder differentiates between positive and negative patches using the InfoNCE loss, allowing it to capture fine-grained quality details. The encoder’s output is then passed to the student’s decoder to guide the model in learning low-level perception and high-level quality prediction.
Furthermore, the student model is continuously refined, with the teacher model being periodically updated through an Exponential Moving Average (EMA) mechanism. This ensures that visual saliency crops are dynamically adjusted, maintaining the stability of local manifold structures during training.

The main contributions of this work are:
\begin{itemize}
\item We propose a novel contrastive learning framework incorporating a local manifold learning strategy. To preserve the local manifold structure, we designate the non-salient regions of an image as intra-class negatives.
\item We introduce the visually salient crops as positive samples, inspired by their strong correlation with human visual perception. These regions play a pivotal role in determining overall image quality.
\item We establish a teacher-student mutual learning paradigm, enabling the dynamic selection of visual saliency crops. This approach ensures the stability of local manifolds throughout the training process.
\end{itemize}


\section{Related Work}
\label{section2}
Since obtaining a reference image is often impractical, NR-IQA~\cite{liu2013single,saad2012blind,zhai2011psychovisual,gu2014using,wang2002no,moorthy2011blind} presents a viable yet challenging alternative, as it directly estimates quality without relying on a reference. The following discussion covers contemporary developments in both NR-IQA methods and the related area of local manifold learning.

\subsection{NR-IQA via Vision Transformer}
Vision Transformers (ViT)~\cite{ViT} have gained prominence in computer vision tasks, including IQA. Hybrid Transformer-based approaches\cite{TReS,TIQA} first extract perceptual features using CNNs before passing through a Transformer encoder. Qin \emph{et al.}\cite{DEIQT} extended this by incorporating Transformer decoders, while Xu \emph{et al.}~\cite{Xu_2024_CVPR} further improved feature fusion by integrating ViT with ResNet.
Despite their strong performance, Transformer-based methods face challenges in effectively representing image quality, as they typically rely on class tokens, originally designed for object recognition. This limitation hinders their ability to accurately model perceptual quality.

\begin{figure*}[t]
\centering{\includegraphics[width=0.8\linewidth]{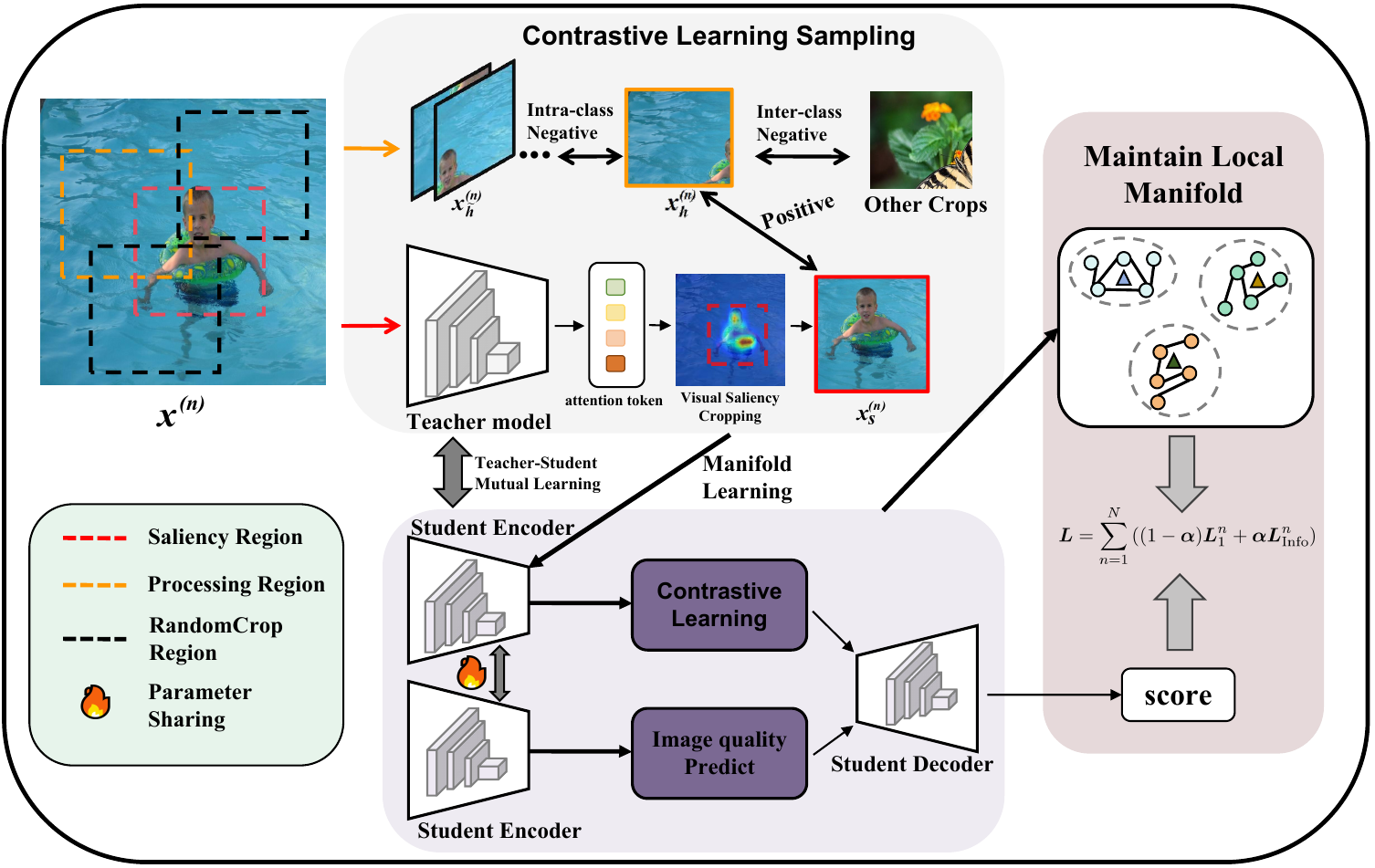}}
	\caption{Overview of the LML-IQA framework.
Given an input image, the teacher model first identifies the most salient region using visual saliency detection. Multiple random crops are generated, with the salient region as the positive sample, random crops from the same image as intra-class negatives, and crops from other images as inter-class negatives. These samples are processed by the student encoder for feature extraction and contrastive learning, followed by quality score prediction with the student decoder. Teacher-student mutual learning is enabled via an exponential moving average (EMA) mechanism.}
    \label{modelStructure}
\end{figure*}
\subsection{NR-IQA based on Contrastive Learning}
Self-supervised contrastive learning has become increasingly popular in NR-IQA research~\cite{madhusudana2022image,saha2023re,zhang2023blind}. CONTRIQUE~\cite{madhusudana2022image} introduced an strategy where images with similar distortion types and levels were grouped into the same class. However, relying on predefined distortion labels limited sample diversity.

To mitigate this, Zhang \emph{et al.}\cite{zhang2023blind} further expanded NR-IQA by linking textual and image information via the CLIP model, improving multi-task generalization. Li\cite{li2024integrating} further captures global contextual knowledge in contrastive learning.

Unlike existing methods, which primarily focus on inter-sample differences, our work emphasizes intra-sample variations. Conventional contrastive learning assumes that different crops of the same image have identical perceptual quality, an assumption that disregards the significance of local regions. Since human visual perception prioritizes salient areas, our approach differentiates between salient and non-salient regions by treating the latter as intra-class negatives. This provides a more fine-grained and robust quality assessment.

\subsection{Manifold Learning in IQA}
Local Manifold Learning (LML)~\cite{ma2010local,chui2018deep}, which aims to discover a low-dimensional manifold within high-dimensional data, was adapted by Jiang \emph{et al.}~\cite{jiang2018perceptual} to reduce the dimensionality of RGB images. Their approach resulted in an improved stereoscopic IQA that was more consistent with human visual perception. Similarly, Guan \emph{et al.}~\cite{guan2018no} utilized manifold learning in High-Dynamic-Range (HDR) image assessment, projecting high-dimensional features onto a lower-dimensional manifold to improve prediction accuracy. However, these approaches predominantly rely on conventional dimensionality reduction techniques, lacking the representational power of deep learning.
In contrast, our method integrates contrastive learning to explicitly model data manifold structures while harnessing deep learning for more expressive and adaptable feature representations.
\section{Method}\label{section3}
\subsection{Overview}
This section elaborates on the proposed LML-IQA. The overview of the LML-IQA framework is provided in Fig.~\ref{modelStructure}. LML-IQA comprises a teacher-student architecture, where the teacher module performs visual saliency-based cropping, generating positive samples for contrastive learning in the student model. Leveraging visually salient regions as positive samples enhances the consistency and representational capability of the learned features.
The student model undertakes two interconnected tasks: contrastive learning and image quality prediction. Both share a common feature extractor for joint optimization. By contrasting salient regions with non-salient ones, the model reinforces its focus on perceptually important areas, which aligns with human attention mechanisms in quality assessment.
Additionally, our approach incorporates the EMA algorithm to facilitate ongoing model refinement. This serves the dual purpose of enhancing the precision of the teacher model’s saliency cropping and providing more consistent positive samples for the contrastive learning task.


\subsection{Local Manifold Learning}
In traditional quality-aware pre-training, all patches from a single image are considered a positive class, set against the negative classes of other images. The primary drawback is a failure to account for feature variations within the same image, as the model solely prioritizes pushing different images apart. This oversight can cause a loss of detailed information (local manifold collapse). Our proposed solution is a new contrastive learning framework, grounded in local manifold learning, designed to address this specific challenge. We define salient regions within an image as positive samples by introducing visual saliency cropping. To preserve local manifold structure, non-salient regions are designated intra-class negatives, while crops from other images serve as inter-class negatives. This setup strengthens the overall manifold differentiation.
Combining positive and negative samples ensures that crops remain sufficiently similar to maintain consistent representation yet distinct enough to prevent local manifold collapse.

\subsubsection{Visual Saliency Cropping}

We argue that the standard practice of using random crops as positive samples in contrastive learning is flawed. These samples are often uninformative and provide little guidance on perceptual quality. This is contrary to human vision, which prioritizes salient regions as the primary basis for determining overall image quality. To align with this principle, we introduce visual saliency cropping. Precisely, we extract positive samples by identifying and cropping an image's most visually significant areas. This process is guided by the teacher model, which leverages a self-attention mechanism to detect and extract these key regions.

We first resize the image to a fixed resolution of ${224\times 224}$ and partition it into $S$ patches. A learnable class token is then prepended to this sequence of patches. This complete set of $S+1$ tokens is subsequently projected by three linear layers to generate the query, key, and value matrices, denoted as ${\boldsymbol Q, \boldsymbol K, \boldsymbol V} \in \mathbb{R}^{(S+1)\times D}$.

With the self-attention mechanism, we derive the attention map  $\mathbf{A}_{t2t} \in \mathbb{R}^{(S+1)\times (S+1)}$, computed as $\mathbf{A}_{t2t} = \mathrm{softmax}(\mathbf{Q}\mathbf{K}^\top/\sqrt{D})$ , which quantifies the attention each token assigns to others. We then extract the class-to-patch attention $\mathbf{A}_{c2p}$ from the $\mathbf{A}_{t2t}$, defined as $\mathbf{A}_{c2p} = \mathbf{A}_{t2t}[0,1:S+1]$.
Since deeper layers capture more discriminative features while shallower layers retain low-level visual details, we integrate token-to-token attention from the last 
$K$ transformer layers to enhance the accuracy of image saliency estimation.

To identify the image's saliency region, we analyze the attention map $\hat{\mathbf{A}}_{c2p}$, which is a matrix of size $\mathbb{R}^{\sqrt{S} \times \sqrt{S}}$. The most salient area is determined by locating the $N \times N$ window within this map that contains the maximum possible sum of attention weights.The procedure is detailed below:
\begin{equation}
    \hat{\mathbf{A}}_{c2p} = \frac{1}{K}\sum_{l}^{l+K}{\mathbf{A}}^{l}_{c2p},
\end{equation}
\begin{equation}
\mathop{\arg\max}\limits_{a,b} = \sum_{n=0}^{M}\sum_{m=0}^{M}{\hat{\mathbf{A}}_{c2p}}[a+n, b+m].
\end{equation}
This region is then designated as the positive sample for contrastive learning.

\subsubsection{Quality-Aware Training}
Within contrastive learning frameworks for image quality assessment, positive pairs are conventionally generated by randomly sampling two patches from a single distorted image. At the same time, negative samples are taken from different distorted images.
However, using randomly cropped patches as positives may not be ideal, as perceptual quality varies across different regions of an image. Human attention is unevenly distributed, often prioritizing central over peripheral areas.
Salient regions better represent overall perceived quality. Thus, we aim to ensure that positive pairs capture quality features that align more closely with the characteristics of these visually significant areas.
Consider $T$ distorted images, represented as $x^1, x^2 ... x^T$. From each source image $x^t$, a collection of samples is generated to enable contrastive learning. This collection is comprised of two components: a set of $H$ randomly cropped views, denoted as $x^t_1, x^t_2, \dots, x^t_H$, and the primary salient region of the image, $x^t_s$.
Within a batch, each patch $x^ t_{i}$ pairs with its relevant salient region $x^ t_{s}$ as a positive sample. Meanwhile, $x^t_{i}$ forms negative pairs with all non-salient regions, encompassing both intra-class and inter-class negatives, as illustrated in Table~\ref{tab:pretext}.

\begin{table}[t]
    \centering
    \footnotesize
    \caption{Defining Positive, Intra-Class Negative, and Inter-Class Negative Pairs in the Pretext Task.}
    \begin{tabular}{c|c|c}
    \toprule
        Positive Pair & \begin{tabular}[c]{@{}c@{}} Intra-class \\ negative pair \end{tabular}  & \begin{tabular}[c]{@{}c@{}} Inter-class \\ negative pair \end{tabular}  \\
    \midrule
        $(\mathbf{x}^{(t)}_{h},\mathbf{x}^{(t)}_{s})$ 
        & $(\mathbf{x}^{(t)}_{h},\mathbf{x}^{(t)}_{\tilde{h}})$
        & $(\mathbf{x}^{(t)}_{h},\mathbf{x}^{(\tilde{t})}_{\ast})$ \\
    \bottomrule
    \end{tabular}
    
    \label{tab:pretext}
\end{table}


To generate features for contrastive training, an input patch $\mathbf{x}^{(t)}_h$ is mapped to a unit-norm embedding $\mathbf{f}^{(t)}_h$ by a Transformer model $\mathcal{F}$, according to the equation:
$
\mathbf{f}^{(t)}_h = \frac{\mathcal{F}(\mathbf{x}^{(t)}_h)}{||\mathcal{F}(\mathbf{x}^{(t)}_h)||_2}
$
The quality similarity between "query" and "key" patches in a batch is subsequently measured using the InfoNCE loss function. Specifically, the loss for an image $x^t$ is expressed as follows:

\begin{equation}
\small
\begin{aligned}
    \mathcal{P}_\mathrm{Intra}^t =  \sum_{\tilde{h}\neq h}^H \exp(\mathbf{f}^{(t)}_{h}\cdot \mathbf{f}^{({t})}_{\tilde{h}}/\tau) ,
\end{aligned}
\end{equation}

\begin{equation}
\small
\begin{aligned}
    \mathcal{P}_\mathrm{Inter}^t = \sum_{\tilde{t}\neq t}^T \sum_{h^{\prime}=1}^H \exp(\mathbf{f}^{(t)}_{h}\cdot \mathbf{f}^{(\tilde{t})}_{h^{\prime}}/\tau) ,
\end{aligned}
\end{equation}

\begin{equation}
\small
\begin{aligned}
    \mathcal{L}_\mathrm{Info}^t  =
    & - \log  
    \frac{\exp(\mathbf{f}^{(t)}_{h}\cdot \mathbf{f}^{(t)}_{s}/\tau)}
    { { \mathcal{P}_\mathrm{Intra}^t } + { \mathcal{P}_\mathrm{Inter}^t } },
    \label{qcLoss}
\end{aligned}
\end{equation}

The quality score for an individual patch $x^t_h$ is predicted by sequentially passing it through a Transformer Encoder $\mathcal{F}$, a Decoder $\mathcal{G}$, and a final MLP layer, yielding $\hat{\text{Y}}^t_h = \text{MLP}(\mathcal{G} (\mathcal{F} (x^t_h)))$. This estimation is part of a broader training process involving both intra-class ($\mathcal{P}_\mathrm{Intra}^t$) and inter-class ($\mathcal{P}_\mathrm{Inter}^t$) interactions. The process is controlled by a temperature value $\tau$, with each batch containing $T$ images from which $H$ patches are cropped.

Next, the $\mathcal{L}_1$ loss is calculated between the predicted quality score $\hat{\text{Y}}^t_h$ and the ground truth $\text{Y}^t_h$:

\begin{equation}
\small
\begin{aligned}
    \mathcal{L}_{1}^t = \sum_{h=1}^H \|  {\,{ {{\hat{\text{Y}}^t_h}}}} -{\text{Y}^t_h\,} \|_1.
\end{aligned}
\end{equation}

\subsubsection{Qualitative Theoretical Justification}~\label{Justification} 
We herein present a theoretical analysis to demonstrate that the loss function based on local manifold learning outperforms conventional contrastive loss. By integrating intra-class negative samples into our framework, we effectively reduce the expected loss, as validated by Eq.~\ref{proof1}.
\begin{equation}
\begin{aligned}
    &\because \mathbb{E}(\lvert y-y_{\varepsilon} \rvert - \lvert y_{\gamma}-y_{\delta} \rvert) \\
    &=\mathbb{E} (\lvert y -(y+\varepsilon_\omega) \rvert  -\lvert (y+ \varepsilon_\gamma) -(y+\varepsilon_\delta) \rvert)\\
    &=\mathbb{E} (\lvert \varepsilon_\omega \rvert  + \lvert \varepsilon_\gamma -\varepsilon_\delta \rvert) \quad  \quad  \text{\footnotesize{($\varepsilon_1 \leq \ldots \leq \varepsilon_\omega \leq \ldots \leq \varepsilon_t$) }}\\
    &= \frac{1}{t} \sum_{\omega} \lvert {\varepsilon_\omega} \rvert - \frac{2}{t(t-1)} \sum_{\omega}(2\omega-t-1) {\varepsilon}_{\omega}  \\
    &=
    \begin{cases}
     \frac{1}{t} \lvert \varepsilon_1 \rvert + \frac{2}{t} \varepsilon_1 \geq - \frac{1}{t} \lvert \varepsilon_1 \rvert, & \omega = 1 \\
    \frac{1}{t} \lvert \varepsilon_t \rvert - \frac{2}{t} \varepsilon_t \geq - \frac{1}{t} \lvert \varepsilon_t \rvert, & \omega = t
    \end{cases}   \\
    &\therefore  \mathbb{E} (\lvert \varepsilon_\omega \rvert  + \lvert \varepsilon_\gamma -\varepsilon_\delta \rvert) \geq - \lvert \varepsilon_\omega \rvert \\
     &\because -\lvert \varepsilon_\omega \rvert \leq \mathbb{E} (\lvert \varepsilon_\omega \rvert  + \lvert \varepsilon_\gamma -\varepsilon_\delta \rvert) \leq  \lvert \varepsilon_\omega \rvert \\
     &\therefore  \lvert \lvert y-y_{\varepsilon} \rvert - \lvert y_{\gamma}-y_{\delta} \rvert \rvert \leq  \lvert \lvert y-y_{\varepsilon} \rvert \rvert \\
\end{aligned}
\label{proof1}
\end{equation}

We define $y_{\varepsilon}$ as the positive class, with $y_{\gamma}$ and $y_{\delta}$ representing inter-class negatives. While our proofs operate on the score level, the method's effectiveness is preserved at the feature level because of the direct mapping outlined in Eq.~\ref{proof2}. This robustness, in turn, suggests a wider applicability for our approach.


\begin{equation}
\begin{aligned}
    y &= \mathcal{G}(f)  \\
    ( y+\Delta y) \iff ( f+& \Delta f ) \quad \text{\footnotesize{ $\lvert \Delta y \rvert \to 0 , \lvert \Delta f \rvert \to 0 $}}\\
    \mathbb{E}(\lvert y-y_{\varepsilon} \rvert - \lvert y_{\gamma}-y_{\delta} \rvert)   &\iff  \mathbb{E}(\lvert f-f_{\varepsilon} \rvert - \lvert f_{\gamma}-f_{\delta} \rvert).
\end{aligned}
\label{proof2}
\end{equation}
The terms in this equation are defined as follows: $f$ represents the feature vector, $y$ is the predicted quality score, and $\mathcal{F}$ and $\mathcal{G}$ are the functions governing the mapping relationship.

\subsubsection{Total Loss for Training} The overall loss comprises a hybrid loss function composed of two key components: a contrastive loss and an $L1$ loss relative to the ground truth. The contrastive loss helps differentiate between various distortions, while the $L1$ loss further guides quality score prediction. The final loss is formulated as follows:

\begin{equation}
\small
\begin{aligned}
    \mathcal{L} = \sum_{n=1}^N ((1-\alpha) \mathcal{L}_{1}^n + \alpha  \mathcal{L}_\mathrm{Info}^n),
\end{aligned}
\label{equ5}
\end{equation}
where $\alpha$ acts as the balancing coefficient that adjusts the relative importance between the dual loss components, while $N$ denotes the batch size corresponding to the total quantity of images processed simultaneously..

\subsection{Teacher-Student Co-mutual Learning}
We propose a dual-stream learning framework in which a teacher model performs visual-saliency-driven cropping and a student model treats the resulting crops as positive pairs for contrastive training. To keep the teacher’s quality awareness evolving—from early attention-only masks to later quality-aware regions—and to supply the student with increasingly consistent positives, we softly synchronize the two models through exponential moving average EMA(Exponential Moving Average). Concretely, the teacher’s parameters are refreshed as:


\begin{equation}
\theta_{t} \leftarrow \beta \theta_{t} + (1-\beta) \theta_{s}.
\label{equ6}
\end{equation}
Let $\theta_{t}$ and $\theta_{s}$ represent the weights of the teacher and student models, respectively, with $\beta$ denoting the EMA coefficient. This co-learning methodology enhances the teacher model's accuracy and stability on the visual saliency cropping task while simultaneously providing superior guidance for the student model's training.
\section{Experiments}\label{section5}
\begin{table*}[]
\normalsize
\caption{A performance comparison is presented using the average SRCC and PLCC metrics on standard IQA datasets. The highest performance is highlighted in \textbf{bold}, while the second-best is \underline{underlined}.}
\renewcommand\arraystretch{1.2}
\centering
\resizebox{1\textwidth}{!}{
\begin{tabular}{l|cc|cc|cc|cc|cc|cc|cc|cc}
\toprule[1.2pt]
\multirow{2}*{\bf{IQA Method}} & \multicolumn{2}{c|}{\bf{CSIQ}} &
\multicolumn{2}{c|}{\bf{LIVE}} &
\multicolumn{2}{c|}{\bf{KADID}} &
\multicolumn{2}{c|}{\bf{TID2013}} &
\multicolumn{2}{c|}{\bf{KonIQ}} &
\multicolumn{2}{c|}{\bf{LIVEC}} &
\multicolumn{2}{c|}{\bf{SPAQ}} &
\multicolumn{2}{c}{\bf{LiveFB}} \\ \cline{2-17}
~ & \bf{SRCC} & \bf{PLCC}
& \bf{SRCC} & \bf{PLCC}
& \bf{SRCC} & \bf{PLCC}
& \bf{SRCC} & \bf{PLCC}
& \bf{SRCC} & \bf{PLCC}
& \bf{SRCC} & \bf{PLCC}
& \bf{SRCC} & \bf{PLCC}
& \bf{SRCC} & \bf{PLCC} \\
\midrule[1pt]

BRISQUE~\cite{BRISQUE} & 0.812 & 0.748 & 0.929 & 0.944 & 0.528 & 0.567 & 0.626 & 0.571 & 0.681 & 0.685 & 0.629 & 0.629 & 0.809 & 0.817 & 0.303 & 0.341 \\
DIIVINE~\cite{saad2012blind} & 0.804 & 0.776 & 0.892 & 0.908 & 0.413 & 0.435 & 0.643 & 0.567 & 0.546 & 0.558 & 0.588 & 0.591 & 0.599 & 0.600 & 0.092 & 0.187 \\

BIECON~\cite{BIECON} & 0.815 & 0.823 & 0.958 & 0.961 & 0.623 & 0.648 & 0.717 & 0.762 & 0.651 & 0.654 & 0.613 & 0.613 & {-} & {-} & 0.407 & 0.428 \\
ILNIQE~\cite{ILNIQE} & 0.822 & 0.865 & 0.902 & 0.906 & 0.534 & 0.558 & 0.521 & 0.648 & 0.523 & 0.537 & 0.508 & 0.508 & 0.713 & 0.712 & 0.294 & 0.332 \\

WaDIQaM~\cite{bosse2017deep} & 0.852 & 0.844 & 0.960 & 0.955 & 0.739 & 0.752 & 0.835 & 0.855 & 0.804 & 0.807 & 0.682 & 0.671 & {-} & {-} & 0.455 & 0.467 \\
MetaIQA~\cite{zhu2020metaiqa} & 0.899 & 0.908 & 0.960 & 0.959 & 0.762 & 0.775 & 0.856 & 0.868 & 0.850 & 0.887 & 0.802 & 0.835 & {-} & {-} & 0.507 & 0.540 \\
DBCNN~\cite{zhang2018blind} & 0.946 & 0.959 & 0.968 & 0.971 & 0.851 & 0.856 & 0.816 & 0.865 & 0.875 & 0.884 & 0.851 & 0.869 & 0.911 & 0.915 & 0.545 & 0.551 \\
HyperIQA~\cite{hypernet} & 0.923 & 0.942 & 0.962 & 0.966 & 0.852 & 0.845 & 0.840 & 0.858 & 0.906 & 0.917 & 0.859 & 0.882 & 0.911 & 0.915 & 0.544 & 0.602 \\

MUSIQ~\cite{ke2021musiq} & 0.871 & 0.893 & 0.940 & 0.911 & 0.875 & 0.872 & 0.773 & 0.815 & 0.916 & 0.928 & 0.702 & 0.746 & 0.918 & 0.921 & {0.566} & {0.661} \\
TIQA~\cite{TIQA} & 0.825 & 0.838 & 0.949 & 0.965 & 0.85 & 0.855 & 0.846 & 0.858 & 0.892 & 0.903 & 0.845 & 0.861 & {-} & {-} & 0.541 & 0.581 \\

DACNN~\cite{pan2022dacnn} & 0.943 & 0.957 & 0.978 & 0.980 & 0.905 & 0.905 & 0.871 & 0.889 & 0.901 & 0.912 & 0.866 & 0.884 & 0.915 & 0.921 & {-} & {-} \\
TReS~\cite{TReS} & 0.922 & 0.942 & 0.969 & 0.968 & 0.859 & 0.858 & 0.863 & 0.883 & 0.915 & 0.928 & 0.846 & 0.877 & {-} & {-} & 0.554 & 0.625 \\
DEIQT~\cite{DEIQT} & 0.946 & {0.963} & {0.980} & {0.982} & 0.889 & 0.887 & \underline{0.892} & \underline{0.908} & {0.921} & {0.934} & {0.875} & {0.894} & {0.919} & {0.923} & {0.571} & {0.663} \\
Re-IQA ~\cite{saha2023re} & {0.947} & 0.960 & 0.970 & 0.971 & 0.872 & 0.885 & 0.804 & 0.861 & 0.914 & 0.923 & 0.840 & 0.854 & {0.918} & {0.925} & - & - \\

LoDa ~\cite{{Xu_2024_CVPR}} & {-} & {-} & {0.975} & {0.979} & \textbf{0.931} & \textbf{0.936} & {0.869} & {0.901} & \textbf{0.932} & \textbf{0.944} & 0.876 & {0.899} & \textbf{0.925} & \textbf{0.928} & \underline{0.578} & \textbf{0.679} \\
QFM-IQM ~\cite{10.5555/3692070.3693183} & \underline{0.954} & \underline{0.965} & \underline{0.981} & \underline{0.983} & 0.906 & 0.906 & {-} & {-} & 0.922 & 0.936 & \underline{0.891} & \underline{0.913} & 0.920 & 0.924 & {0.567} & {0.667} \\
\midrule
std & ±0.005& ±0.003 &±0.003 &±0.003 &±0.005 &±0.005 & ±0.014 & ±0.011 & ±0.0002 & ±0.001 & ±0.013 & ±0.006& ±0.002 & ±0.001 & ±0.001 & ±0.011 \\
LML-IQA (Ours) & \textbf{0.968} & \textbf{0.977} & \textbf{0.982} & \textbf{0.984} & \underline{0.915} & \underline{0.918} & \textbf{0.931} & \textbf{0.942} & \underline{0.923} & \underline{0.938} & \textbf{0.891} & \textbf{0.914} & \underline{0.921} & \underline{0.926} & \textbf{0.585} & \underline{0.673} \\

\bottomrule[1.5pt]
\end{tabular}
}
\label{tab:all}
\end{table*}
\subsection{Experimental Datasets and Evaluation Metrics}
The LML-IQA model is evaluated on eight standard NR-IQA datasets, comprising four synthetic and four real-world sets. The synthetic datasets are LIVE~\cite{sheikh2006statistical}, CSIQ~\cite{larson2010most}, TID2013~\cite{ponomarenko2015image}, and KADID~\cite{lin2019kadid}, while the real-world ones include LIVEC~\cite{ghadiyaram2015massive}, KonIQ~\cite{hosu2020koniq}, LiveFB~\cite{ying2020patches}, and SPAQ~\cite{fang2020Perceptual}. The synthetic datasets are created by applying various distortion types, such as Gaussian blur and random noise, to a finite set of original, pristine images.
In contrast, the real-world dataset consists of images collected from multiple photographers using various mobile devices. This methodology ensures the capture of authentic artifacts that arise in natural settings.

To evaluate the predictive performance and monotonicity of the LML-IQA model, we utilize two commonly used standard evaluation metrics: the Spearman Rank Correlation Coefficient (SRCC) and the Pearson Linear Correlation Coefficient (PLCC). These metrics range from $0$ to $1$, with higher values signifying stronger predictive ability.

\subsection{Implementation Details}
We adhere to a standard training protocol. Input images are first cropped into eleven $224 \times 224$ patches (ten random, one saliency-based). Our architecture utilizes the ViT-S Transformer encoder from DeiT~III~\cite{touvron2022deit} and a 1-layer Transformer Decoder. The attention score is computed as the average of the last three encoder layers.

We train the model for a total of 9 epochs, setting the initial learning rate to $2 \times 10^{-4}$. A decay factor of 10 is applied to the learning rate every 3 epochs. The batch size is scaled according to the dataset size. For validation, we perform ten independent runs, each time randomly splitting the data into an 80\% training and 20\% testing set. The final reported metrics are the average PLCC and SRCC from these runs.

\subsection{Overall Performance Comparison}

The performance of LML-IQA is summarized across eight datasets in Table~\ref{tab:all}, demonstrating its consistently competitive capabilities. Notably, on the CSIQ and TID2023 datasets, LML-IQA shows a substantial leap in performance. When compared to the second-best method, our model yields SRCC gains of \textbf{0.014} and \textbf{0.039}, respectively. This enhanced performance is a direct result of incorporating visual saliency cropping and intra-class negative classes, which work to heighten the model's perception of quality variations and prevent the collapse of the manifold structure. Considering the challenge posed by the diversity of image content and distortion types in these benchmarks, these results strongly confirm the effectiveness of LML-IQA for accurate image quality characterization.

   

\subsection{Generalization Capability Validation}
\noindent \textbf{Cross-Dataset Performance.}We further investigated the generalization performance of LML-IQA through cross-dataset validation experiments. The protocol involved training our model on one dataset and directly testing it on another, with no fine-tuning or adaptation of parameters. To maintain simplicity and ensure universality, several such experiments were carried out. Table \ref{tab:cross} displays the outcomes, reported as the average SRCC values. LML-IQA emerged as the best-performing method in four of these experiments. Impressively, some of the cross-dataset performance metrics were superior to those of the supervised approaches listed in Table~\ref{tab:all}.

By employing contrastive learning, we guide the model to capture fundamental features indicative of image quality, preventing it from merely memorizing the labels of the training set. The model utilizes comparisons within and between classes to preserve the local manifold while simultaneously constructing a generalized manifold structure that is robust.

\begin{table}[t]
\caption{\textbf{SRCC for Cross-Dataset Validation.} The best and second-best results are highlighted in \textbf{bold} and with an \underline{underline}, respectively.}
\centering
\footnotesize
{
\begin{tabular}{lccccc}
\toprule[1.5pt]
Training & LIVEC & { LIVEFB } & KonIQ & CSIQ & LIVE \\
\midrule[0.25pt]
Testing & KonIQ & LIVEC & LIVEC & LIVE & CSIQ \\
\midrule[1pt]
P2P-BM~\cite{ying2020patches} & 0.740 & 0.738 & 0.770 & - & 0.712 \\
TReS~\cite{TReS} & 0.733 & 0.740 & 0.786 & - & 0.761 \\
DEIQT\cite{DEIQT} & 0.744 & \underline{0.781} & {0.794} & \underline{0.932} & \underline{0.781} \\
LoDa~\cite{Xu_2024_CVPR} & \underline{0.745} & \textbf{0.805} & \underline{0.811} & {-} & {-} \\
\midrule[1pt]
LML-IQA(Ours) & \textbf{0.756} & {0.763} & \textbf{0.818} & \textbf{0.951} & \textbf{0.833} \\
\bottomrule[1.5pt]
\end{tabular}
}
\label{tab:cross}
\end{table}

\begin{figure*}[t]
\centering{\includegraphics[width=0.8\textwidth]{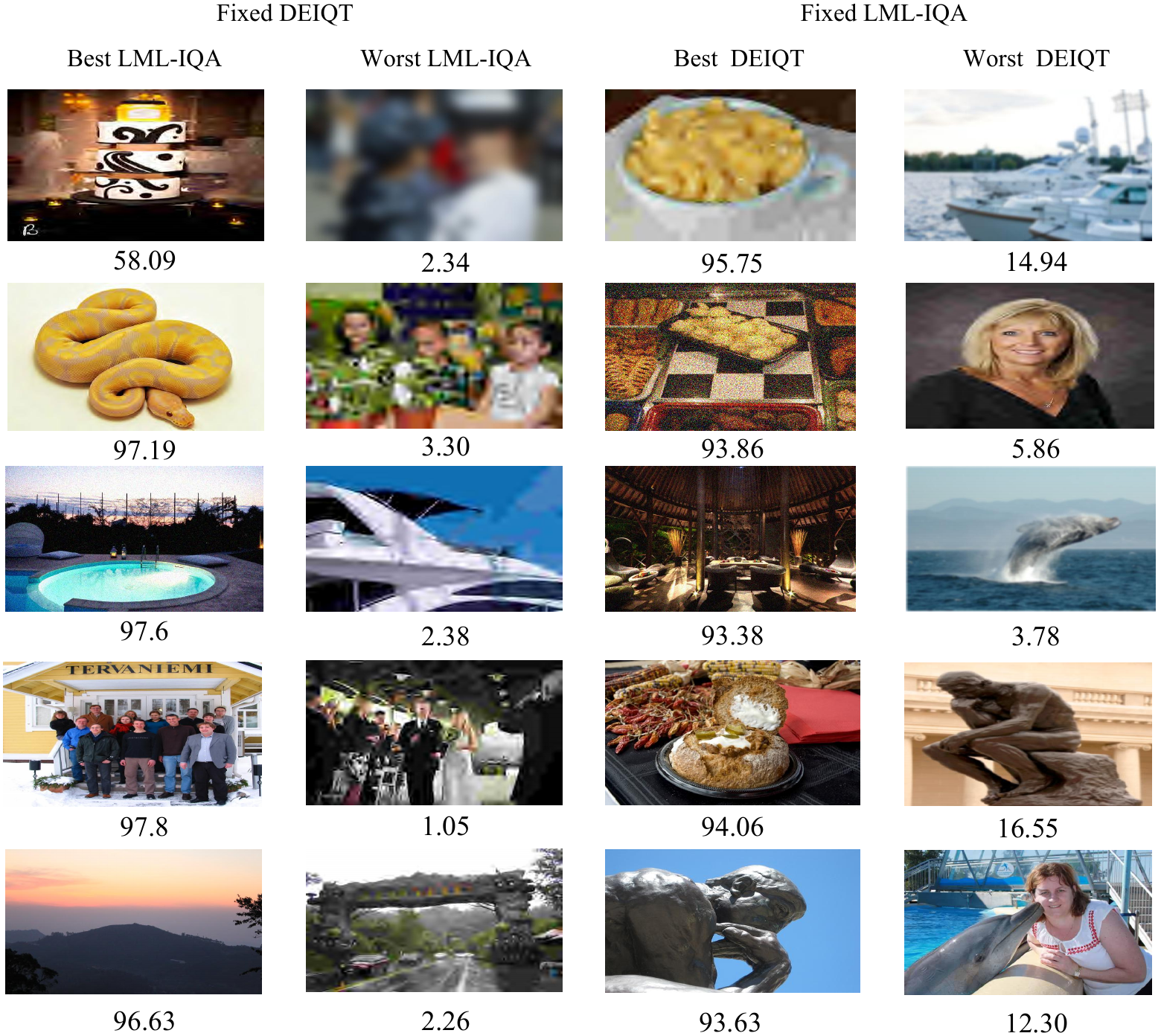}}
\vspace{-8pt}
	\caption{gMAD competition results between DEIQT~\cite{DEIQT} and LML-IQA. The first two columns represent LML-IQA as the attacker and DEIQT as the defender, while the roles are reversed in the last two columns. Each row, from top to bottom, fixs the defender at different quality level constant, ranging from low to high. The numerical values below each image indicate the attacker's perceived quality score.
 }
 \label{gmad}

\end{figure*}
\noindent \textbf{gMAD Testing.}
For an additional assessment of generalization, we trained our models on the entire LIVE database and evaluated their performance on the Waterloo Database~\cite{ma2016waterloo} under the gMAD competition~\cite{ma2016group} guidelines. The gMAD competition strategically selects image pairs that exhibit the most significant quality difference according to an attacking IQA model, to challenge a defending model that perceives these pairs as having similar quality. These pairs are then shown to observers, who judge the robustness of the attacker and defender. As shown in Fig.~\ref{gmad}, the first two columns illustrate instances where our model acts as the attacker against the DEIQT method, with each column corresponding to images predicted by the defender to be of lower and higher quality, respectively. In the final two columns, our model is positioned as the defender, with image pairs representing lower and higher quality levels.
The results depicted in Fig.~\ref{gmad} demonstrate that when our model takes on the role of the attacker, it effectively identifies quality differences in image pairs that the defender perceives as having consistent quality, assigning significantly different scores. 
Conversely, when our model serves as the defender, the attacker selects image pairs with minor perceived quality differences but predicts significantly divergent scores.
This highlights LML-IQA's outstanding offensive and defensive capabilities against DEIQT.

Additionally, our model successfully identified images with overall decent quality but actually degraded by white noise (second row, third column), even though these images successfully misled the DEIQT into assessing them as high-quality, our results further confirm the proposed model's robust generalization ability and its proficiency in handling complex distortions found in authentic images.

\begin{table}[t]
 \normalsize
 \vspace{-8pt}
 \caption{Ablation experiments on TID and LIVEC datasets. \textbf{Bold} entries indicate the best performance.}
\renewcommand\arraystretch{1.2}
    \centering
      \resizebox{0.5\textwidth}{!}{
        \begin{tabular}{l|cc|cc}
        \toprule[1.5pt]
        \multirow{2}*{{Module}}
        &  \multicolumn{2}{c|}{TID} & \multicolumn{2}{c}{LIVEC} \\ \cline{2-5} 
         & SRCC & PLCC & SRCC & PLCC \\ \midrule[1pt]
        Baseline  & 0.873 & 0.896 & 0.870 & 0.887 \\
        \midrule[0.25pt]
        $\textbf{+}$ Random Crop  & 0.885 & 0.904 & 0.874 & 0.894 \\
         $\textbf{+}$ Visual Saliency Crop  & 0.902 & 0.918 & 0.883 & 0.903 \\
        $\textbf{+}$ Mutual Learning  & 0.920 & 0.934 & 0.884 & 0.910 \\
        \midrule[0.25pt]
        LML-IQA & \textbf{0.931} & \textbf{0.942} & \textbf{0.891} & \textbf{0.914} \\
        \bottomrule[1.5pt]
        \end{tabular}
        }
   
    \label{tab:abla}
\end{table}

\begin{table}[t]
\caption{Ablation Study on the Parameter M for the Visual S saliency Cropping Module.} 
\setlength\tabcolsep{12pt}
  \centering
 
  {
    \begin{tabular}{l|cc|cc}
    \toprule[1.25pt]
    \multirow{2}*{$M \times M$}
    &  \multicolumn{2}{c|}{CSIQ} & \multicolumn{2}{c}{LIVEC}  \\
    \cline{2-5}      & \multicolumn{1}{c}{SRCC} & \multicolumn{1}{c|}{PLCC} & \multicolumn{1}{c}{SRCC} & \multicolumn{1}{c}{PLCC} \\
    \midrule
    {$5 \times 5$}  
    & {0.966} & {0.975} & {0.885} & {0.910}\\
    {$6 \times 6$}  
    & {0.965} & {0.974} & {0.885} & {0.906}\\
   {$7 \times 7$}   
    &  0.965 & 0.974 & 0.884 & 0.906
    \\
   {$8 \times 8$}    
    &  \textbf{0.968} & \textbf{0.977} & \textbf{0.891} & \textbf{0.914}
    \\
   {$9 \times 9$}  
    &  0.965 & 0.974 & 0.885 & 0.907
    \\
    {$10 \times 10$}  
    &  0.963 & 0.973 & {0.880} & {0.905}
    \\
    \bottomrule[1.25pt]
    \end{tabular}%
    }
 
  \label{table5}
\end{table}%


\subsection{Ablation Study}
LML-IQA is a novel approach founded on local manifold learning, integrating two core modules: visual saliency cropping and a teacher-student co-learning framework. As detailed in Table~\ref{tab:abla}, we conducted an ablation study to systematically dissect the contribution of each module to the model's overall performance.

\noindent \textbf{Effectiveness of Random vs. Visual Saliency Cropping.}
We first established a baseline using contrastive learning with random cropping, where two crops from the same image form a positive pair and crops from different images serve as negative pairs. The results indicated that contrastive learning alone is beneficial for learning more fundamental image quality representations. Subsequently, replacing random cropping with our proposed visual saliency cropping yielded a significant performance enhancement. This outcome validates our core motivation: that the perception of image quality is primarily driven by salient regions. By aligning the crops with these key areas, we strengthen the similarity within positive pairs, offering more accurate guidance for the contrastive learning process.

Prior to the commencement of training in this particular experiment, the teacher model pre-determines a fixed set of visual saliency regions for the positive pairs.

\noindent \textbf{Effectiveness of Mutual Learning and EMA.}
To enable the dynamic adjustment of these saliency regions, we introduced a teacher-student mutual learning paradigm. In this setup, the teacher generates saliency crops before each training epoch, and the student's weights are used to update the teacher after each epoch. Unlike a 'frozen' teacher with fixed saliency maps, this dynamic update allows for the real-time refinement of salient areas. The student's guidance helps the teacher's focus evolve from initial salient objects to the regions that are truly critical for overall image quality perception.

Additionally, we utilize the Exponential Moving Average (EMA) algorithm to stabilize the teacher's weight updates. This prevents the rapid changes that could otherwise cause inconsistencies in epoch-to-epoch saliency maps and negatively impact the contrastive learning task. This delayed update mechanism ensures that the feature samples being compared remain more consistent, while also helping the model to better concentrate on the most relevant salient areas.

In summary, the ablation experiments confirm that each component significantly contributes to the overall performance of LML-IQA. The proposed visual saliency cropping and teacher-student mutual learning framework provide substantial improvements in both model accuracy and stability, and their synergy advances the development of deep learning models for image quality assessment.

\subsection{Analysis of the cropping size M in Visual Saliency Cropping}

We investigated the influence of the visual saliency patch dimension, $M \times M$, on model performance using the synthetic CSIQ and real-world LIVEC datasets, with $M$ varying from 5 to 10. As shown in Table~\ref{table5}, the performance demonstrated high stability, with fluctuations of less than 1\% across the different values of $M$. An optimal crop size of $8 \times 8$ was determined for both datasets. This is because an increase in patch size provides the model with more reference cues within salient regions, which benefits the contrastive learning mechanism. However, expanding the patch beyond this optimal size introduces irrelevant information and dilutes saliency-specific features, potentially degrading the model's overall performance.


\begin{figure*}[h]
	\centering{\includegraphics[width=1.0\textwidth]{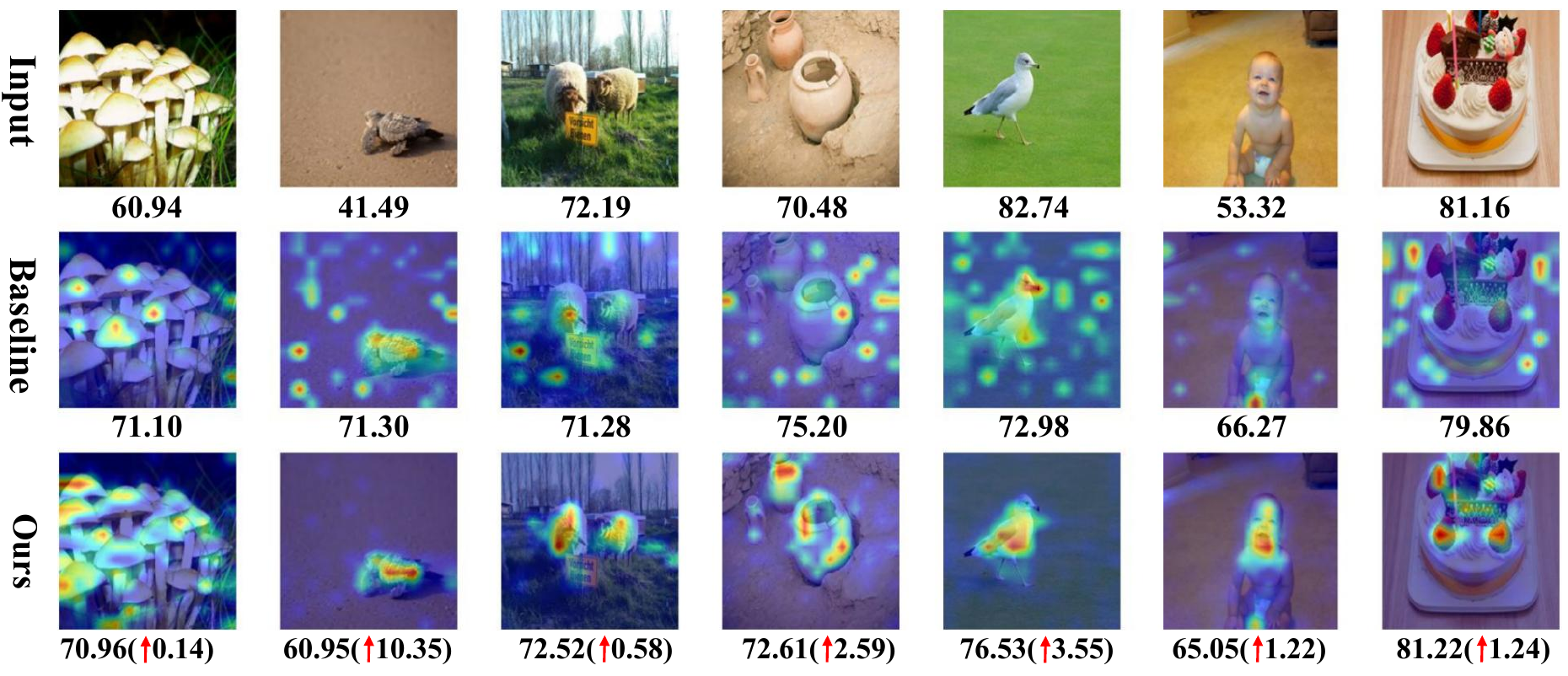}}
	\caption{
    Comparison of Grad-CAM activation maps between the baseline~\cite{DEIQT} and LML-IQA. For each example, the figure displays three rows: (1) the input image, (2) the CAM from the baseline, and (3) the CAM from LML-IQA. The numerical values provided below each column correspond to the Ground Truth score, the baseline's prediction, and our model's prediction, respectively.
 }
 \label{visualize}
\end{figure*}

\begin{table}[t]
 \caption{Assessing the model's data efficiency by training on reduced subsets of the data (20\%, 40\%, and 60\% of the original training set). The top-performing results are highlighted in bold.}
\setlength\tabcolsep{2.5pt}
    \centering
     \footnotesize
     {
        \begin{tabular}{ll|cc|cc|cc}
        \toprule[1.5pt]
         \multirow{2}*{{Training set}}& \multirow{2}*{{Methods}} & \multicolumn{2}{c|}{LIVEC} & \multicolumn{2}{c|}{LIVE} & \multicolumn{2}{c}{KonIQ} \\ \cline{3-8} 
         &  & SRCC & PLCC & SRCC & PLCC & SRCC & PLCC \\ \midrule[1pt]
        \multirow{4}{*}{ 20\%}  & ViT-BIQA & 0.622 & 0.641 & 0.894 & 0.828 & 0.825 & 0.855 \\
         & HyperNet  & 0.776 & 0.809 & 0.951 & 0.950 & 0.869 & 0.873 \\
         & DEIQT & {0.792} & {0.822} & {0.965} & {0.968} & {0.888} & {0.908} \\ 
         & LML-IQA & \textbf{0.811} & \textbf{0.841} & \textbf{0.974} & \textbf{0.976} & \textbf{0.898} & \textbf{0.917} \\ \midrule[0.25pt]
         \multirow{4}{*}{ 40\%} & ViT-BIQA & 0.684 & 0.714 & 0.903 & 0.847 & 0.880 & 0.901 \\
         & HyperNet  & 0.832 & 0.849 & 0.959 & 0.961 & 0.892 & 0.908 \\
         & DEIQT & {0.838} & {0.855} & {0.971} & {0.973} & {0.903} & {0.922} \\ 
         & LML-IQA & \textbf{0.852} & \textbf{0.878} & \textbf{0.977} & \textbf{0.979} & \textbf{0.911} & \textbf{0.929} \\ \midrule[0.25pt]
         \multirow{4}{*}{ 60\%} & ViT-BIQA & 0.705 & 0.739 & 0.915 & 0.856 & 0.903 & 0.916 \\
         & HyperNet  & 0.843 & 0.862 & 0.960 & 0.963 & 0.901 & 0.914 \\
         & DEIQT & {0.848} & {0.877} & {0.972} & {0.974} & {0.914} & {0.931} \\ 
         & LML-IQA & \textbf{0.877} & \textbf{0.898} & \textbf{0.979} & \textbf{0.981} & \textbf{0.915} & \textbf{0.931} \\  \bottomrule[1.5pt]
        \end{tabular}
        }
   
    \label{tab:eff}
\end{table}
\subsection{Data-Efficient Learning Validation}
Data-efficient learning is a crucial objective in IQA, given the expensive nature of image annotation and model training.
Through local manifold learning, LML-IQA is designed to adequately discern perceptual differences in image quality with a reduced amount of data, thereby decreasing the need for large training sets.

To demonstrate this capability, we conducted an experiment by varying the training data size in increments of 20\%, from 20\% to 60\%.
This procedure was repeated ten times for each increment to ensure statistical reliability, with the average performance recorded.
In every trial, a distinct 20\% of the data was reserved for testing.
As presented in Table~\ref{tab:eff}, our approach consistently outperforms other state-of-the-art NR-IQA methods when using an equivalent amount of training data.
It is noteworthy that LML-IQA delivers competitive performance on the LIVE and CSIQ datasets (Table~\ref{tab:all}) with only 40\% of the training data.
At a 60\% training data volume, LML-IQA surpasses all competing methods.

In summary, the method's strength in data-efficient learning comes from its strategy of using saliency cropping to focus on areas highly correlated with image quality and then performing local manifold learning on these regions to master the extraction of image-quality information.

\begin{figure}[t]
\centering{\includegraphics[width=0.49\textwidth]{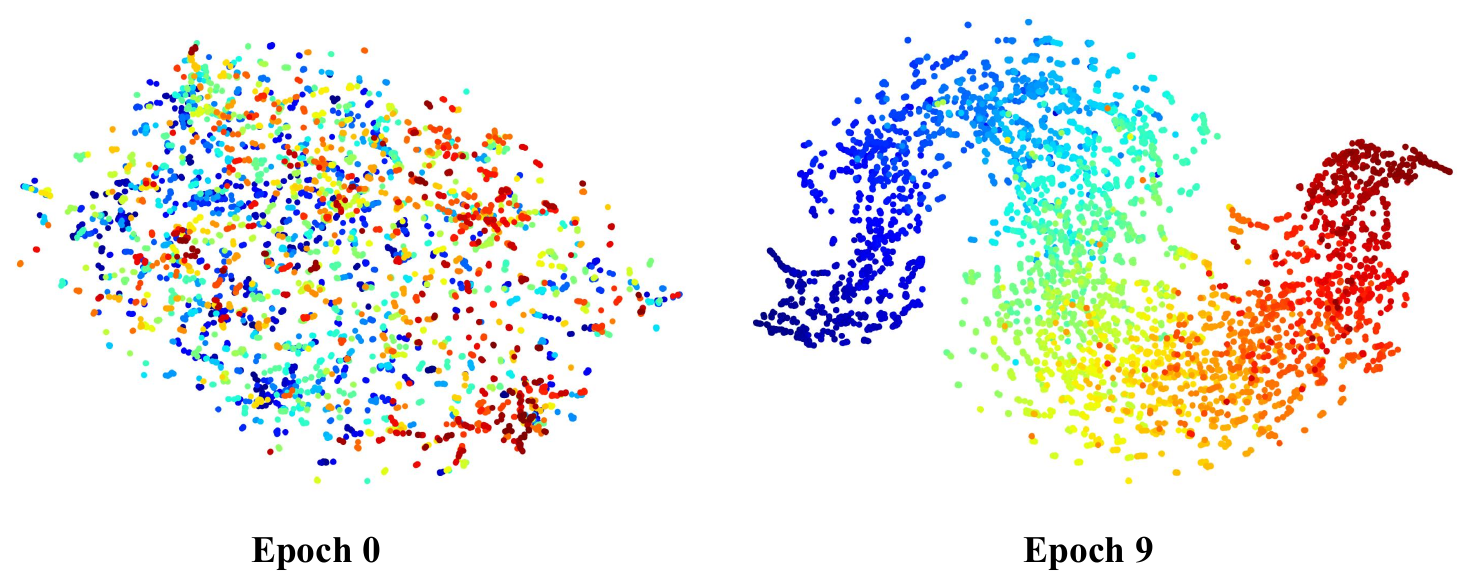}}
	\caption{The t-SNE visualization of the quality features of the LIVEC training set learned by our LML-IQA, with different colors representing different quality score ranges.}
 \label{tsne}
\end{figure}

\subsection{Visualization}
\subsubsection{Visual Analysis of the Quality Attention Mechanism}
We utilize GradCAM~\cite{grad} to generate feature attention maps, as illustrated in Fig.~\ref{visualize}. The visualizations reveal that our model allocates its attention comprehensively and precisely to regions containing significant visual distortions. In contrast, the baseline model tends to focus on irrelevant areas. This distinction demonstrates our model's superior ability to capture correct semantic structures and achieve a more accurate perception of quality. This capability is a direct result of our training strategy, which deliberately emphasizes features from visually salient regions. Moreover, the predicted quality scores further support this observation, showing our model's predictions align more closely with the ground truth compared to the baseline. The effectiveness of our proposed method is strongly corroborated by the cumulative evidence from these results.

\subsubsection{Quality Features Visualization}
To further analyze our approach, we utilized t-SNE~\cite{t-sne} to visualize the feature distribution generated by LML-IQA on the LIVEC dataset. In this visualization, a color gradient represents image quality on a 0-100 scale, where blue dots correspond to lower-quality images with scores below 30 and red dots signify higher-quality images with scores above 80. As illustrated in Fig.~\ref{tsne}, the feature embeddings are initially intermingled across different quality categories. However, after applying local manifold learning, our model displays a remarkable ability to differentiate between low and high quality images. This is demonstrated by the resulting feature space, which exhibits compact intra-class distances that create tight clusters for images of similar quality, and well-separated inter-class distances that ensure clear separation between images of varying quality levels.

\section{Conclusion}\label{section4}
This paper proposes LML-IQA, a novel No-Reference Image Quality Assessment (NR-IQA) model that leverages local manifold learning. Our approach hinges on two key strategies: visual saliency cropping and a novel negative sampling technique. Our approach employs intra-class negative samples to ensure representation uniformity and prevent the local manifold from collapsing. Concurrently, inter-class negative samples are used to structure the global manifold, which enhances the robustness of the final assessment. Additionally, we implement a teacher-student mutual learning scheme using the EMA algorithm. This allows the visual saliency cropping to adapt dynamically while ensuring the stability of the local manifolds. Experimental evaluations on eight IQA benchmarks confirm the effectiveness and superiority of our proposed method.



\bibliographystyle{unsrt}
\bibliography{ref}

\vfill

\end{document}